\documentclass{sig-alternate-05-2015}
\usepackage{amssymb}
\usepackage{graphicx}
\usepackage[ruled,vlined,linesnumbered,algo2e]{algorithm2e}
\usepackage{sidecap}
\usepackage{amsmath}
\usepackage{longtable}
\usepackage{numprint}
\usepackage{url}
\usepackage{xspace}
\usepackage{multirow}
\usepackage{enumitem}
\usepackage[tight,footnotesize]{subfigure}
\usepackage{mdwlist}
\usepackage{balance}

\usepackage{todonotes}
\usepackage{customizedmacros}
\usepackage{workinprogressmacros}
\usepackage{rotating}

\newfont{\mycrnotice}{ptmr8t at 7pt}
\newfont{\myconfname}{ptmri8t at 7pt}
\let\crnotice\mycrnotice%
\let\confname\myconfname%

\newcommand{\approach}{TIEmb}

\begin{document}
\CopyrightYear{2017} 
\setcopyright{acmcopyright} 
\conferenceinfo{WI '17,}{August 23-26, 2017, Leipzig, Germany}
\isbn{978-1-4503-4951-2/17/08}\acmPrice{\$15.00}
\doi{http://dx.doi.org/10.1145/3106426.3106465}

\title{Large-scale Taxonomy Induction Using\\ Entity and Word Embeddings}


\numberofauthors{1} 
\author{\alignauthor Petar Ristoski, Stefano Faralli, Simone Paolo Ponzetto and Heiko Paulheim \\
\affaddr{Data and Web Science Group} \\ \affaddr{University of Mannheim, Germany} \\
\email{\{petar.ristoski,stefano,simone,heiko\}@informatik.uni-mannheim.de}
}

\maketitle

\begin{abstract}
Taxonomies are an important ingredient of knowledge organization, and serve as a backbone for more sophisticated knowledge representations in intelligent systems, such as formal ontologies.
However, building taxonomies manually is a costly endeavor, and hence, automatic methods for taxonomy induction are a good alternative to build large-scale taxonomies.
In this paper, we propose \emph{\approach}, an approach for automatic unsupervised class subsumption axiom extraction from knowledge bases using entity and text embeddings. 
We apply the approach on the WebIsA database, a database of subsumption relations extracted from the large portion of the World Wide Web, to extract class hierarchies in the Person and Place domain.
\end{abstract}


\section{Introduction}
\label{sec:intro}
Semantic ontologies and hierarchies are established tools to represent domain-specific knowledge with dozens of scientific, industrial and social applications \cite{Glass:1995} and, in knowledge bases, the natural way to structure the knowledge. A basic building block of an ontology is a class, where the classes are organized with ``is-a'' relations, or class subsumption axioms (e.g., each \emph{city} is a \emph{place}). It is crucial to define a high quality class hierarchy for a knowledge base in order to allow effective access to the knowledge base from various Natural Language Processing, Information Retrieval, and any Artificial Intelligence systems and tools.

However, manually curating a class hierarchy for a given knowledge graph is time consuming and requires a high cost. For example the DBpedia Ontology \cite{LehmannDBpedia}, which is a central hub for many applications in the Semantic Web domain \cite{SchmachtenbergLOD}, has been manually created based on the most commonly used infoboxes within Wikipedia.
Many recent studies propose automatic extraction of class hierarchies \cite{gupta2016domain,KozarevaetHovy:2010,Velardietal:2013}. The importance of automatic approaches for the induction class hierarchy becomes more apparent when we deal with large scale automatically acquired knowledge bases such as the WebIsA database (WebIsADb) \cite{seitner2016large}.

The WebIsADb is a large collection of more than 400 million hypernymy relations. Relations are extracted from the CommonCrawl,\footnote{\url{http://commoncrawl.org/}} by means of Hearst-like patterns \cite{hearst1992automatic}. Being extracted from raw text from the very diverse Web sources, and using heuristics of varying reliability, the WebIsaDB provides a good coverage, but rather low precision, and cannot be applied as a taxonomy as is.

In recent years, word embedding models have been used heavily in many NLP applications. Such approaches take advantage of the word order in text documents, explicitly modeling the assumption that closer words in the word sequence are statistically more dependent. In the resulting semantic vector space, similar words appear close to each other, and simple arithmetic operations can be executed on the resulting vectors. One of the widely used approach is the word2vec neural language model \cite{mikolov2013efficient,mikolov2013distributed}. Word2vec is a particularly computationally-efficient two-layer neural net model for learning word embeddings from raw text.
Another widely used approach is GloVe \cite{pennington2014glove}, which in comparison to word2vec is not a predictive model, but a count-based model, which learns word vectors by doing dimensionality reduction on a co-occurrence counts matrix. The studies suggest that both models show comparable performances on many NLP tasks \cite{levy2015improving}.

Such word embedding approaches have been adapted for knowledge base embeddings, or entity embeddings. For example, the RDF2vec approach \cite{ristoski2016rdf2vec} is able to embed large knowledge bases, like DBpedia. The approach first performs graph transformations on the complete knowledge base, and then learns entity embeddings using neural language models.

In this paper, we propose the \emph{\approach}~approach for automatic unsupervised class subsumption axiom extraction from knowledge bases using entity and text embeddings. The underlying assumptions behind our approach are: (i) the majority of all instances of the same class are positioned close to each other in the embedded space; (ii) each class in the knowledge base can be represented as a cluster of the instances in the embedded space, defined with a centroid and an average radius; (iii) clusters that completely or partially subsume each other, indicate class subsumption axiom.

\begin{figure}[t]
    \centering
      \frame{\includegraphics[width=0.5\textwidth]{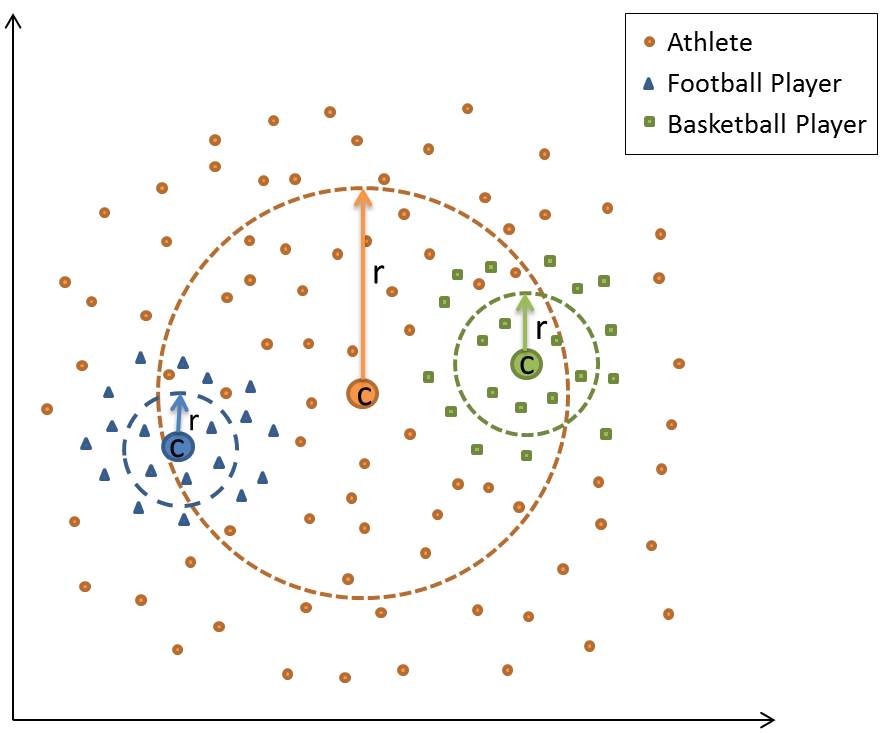}}
  \caption{Example of three classes in two-dimensional feature space}
  \label{fig:exampleClustering}
\end{figure}

Figure \ref{fig:exampleClustering} shows an example of three class clusters projected into a two dimensional feature space. Each of the classes is represented with the class instances, centroid and a radius. As we can observe, the centroids of the ``Football Player'' and ``Basketball Player'' classes are within the radius of the ``Athlete'' class, which indicates that the ``Football Player'' and ``Basketball Player'' are subclasses of the ``Athlete'' class.

The contributions of this paper are the following ones:

\begin{itemize}[leftmargin=4mm]

\item We present a novel unsupervised approach to induce class subsumption axioms using entity and word embeddings.
\item We show that such approach can be applied on large knowledge bases, like DBpedia and the WebIsADb.
\item Finally, we provide the resulting class subsumption axioms in the \textit{Person} and \textit{Place} domains extracted from the WebIsADb. This goes in the direction of semantifying the WebIsADb, and the CommonCrawl in general.

\end{itemize}

The rest of this paper is structured as follows: in section~\ref{sec:related}, we given an overview of related work. We present our approach in section~\ref{sec:approach}, followed by two evaluations in section~\ref{sec:experiments}, one being on DBpedia, one on the WebIsaDB. We conclude with a summary and an outlook on future work.

\section{Related Work}
\label{sec:related}


The backbone of ontologies typically consist of hierarchy of hypernymy relations, namely ``IsA'' relations, typically pairs of the kind $(t,h)$ where $t$ is a concept/term and $h$ one of its generalizations. Hence, in the construction of knowledge bases, the induction of taxonomies represents an intermediate fundamental step. However, manually constructing taxonomies is a very demanding task, requiring a large amount of time and effort. A quite recent challenge, referred to as ontology learning, consists of automatically or semi-automatically creating a lexicalized ontology using textual data from corpora or the Web (see~\cite{Biemann:05,IJID120984} for a survey on the task). As a result, the heavy requirements of manual ontology construction are drastically reduced.


The task of lexicalized taxonomy induction can be started with the extraction of domain specific hypernymy relations from texts.
To this end,~\cite{KozarevaetHovy:2010}~use Hearst-like patterns~\cite{hearst1992automatic} to bootstrap the extraction of terminological sisters terms and hypernyms. Instead, in ~\cite{Velardietal:2013} the extraction of hypernymy relations is performed with a classifier, which is trained on a set of manually annotated definitions from Wikipedia, being able to detect definitional sentences and to extract the definiendum and the hypernym. In the above mentioned systems, the harvested hypernymy relations are then arranged into a taxonomy structure.

In general, all such lexical-based approaches suffer from the limitation of not being sense-aware, which results in spurious taxonomic structures.  To cope with such limitations, in \cite{Farallietal:2016} the authors adopt a corpus-based unsupervised distributional semantics method to harvest fully disambiguated sense inventories, as well as a new approach to clean distributional semantics-based acquired knowledge graphs  \cite{Farallietal:2017}.

In all the above web-based approaches, a problem arises when the systems open to the Web. In fact, in the last year the majority of the Web search engines do not allow to programmatically query the Web. The WebIsADb  \cite{seitner2016large} addresses the ``unavailability'' of the Web indices, providing a large database (more than 400M tuples) of hypernymy relations extracted from the CommonCrawl web corpus.\footnote{\url{https://commoncrawl.org}}

Another group of approaches - instead of inducing a taxonomy from scratch - involve statistical evidence to induce structured hierarchies on existing data sources. 
In~\cite{volker2011statistical} (among others) a schema is statistically induced from the large amount of RDF triples on the Web. Enabling suitable schemas for all those application where logical inference is required.

Recently, word embeddings representations are involved in the task of knowledge hierarchical organization  \cite{fu2014learning}, \cite{zafar2016using}, \cite{gupta2016domain}. However, these methods are only considering hypernym-hyponym relationship extraction between lexical terms, using word embeddings. Instead, in our approach we focus on extracting class subsumption axioms, where each class is represented with a set instances.

\section{Approach}
\label{sec:approach}
Our approach makes use of vector space embeddings. Those are projections of each instance -- e.g., a word or an entity in a graph -- into a low dimensional vector space. The core assumption that we exploit in our approach is that two similar entities are positioned close to each other in that vector space. 

Following that intuition, we can assume that entities which belong to the same class are positioned close to each other in the vector space, since they share some commonalities. Furthermore, instances of a more specific class should be positioned closer to each other on average than instances of a broader class. For example, the class of basketball players is more specific than the class of athletes, so that we assume that basketball players are on average closer to each other in the vector space than athletes.

Furthermore, since basketball players are a subclass of athletes, we assume that the majority of vector space points corresponding to basketball players will be positioned inside the cluster formed by the athletes in general, hence, their centroids will be close to each other as well.

The \approach~approach builds upon those two assumptions. Its pseudocode is shown in Algorithm \ref{algo:clusteringAlgorithm}. The algorithm has two inputs: (1) a knowledge base as its input, which contains a set of instances, where each instance has one or more class types, and (2) the knowledge base embeddings, where each instance is represented as n-dimensional feature vector. The output of the algorithm is a set of class subsumption axioms.

In the first step, the algorithm calculates the centroid and radius for all the classes in the knowledge base (lines 2-8). To do so, we select all instances for all the classes, where each instance is represented with the corresponding embedding vector $v_i$. The centroid of a class represents a vector, which is calculated as the average of the vectors of all the instances that belong to the class (line 6). The radius of the class is calculated as the average distance of all instances in the class to the centroid (line 7).

In the next step, the class subsumption axioms are extracted (lines 9-24). First, for each class $c_1$ in the knowledge base we generate a set of candidate axioms. The algorithm iterates over all pairs of classes, and checks for two conditions before it creates new candidate axiom: for the classes $c_1$ and $c_2$, if the distance  between the centroids of the classes, $distance_{c_1c_2}$, is smaller than the radius of $c_2$ and the radius of $c_1$ is smaller than the radius of $c_2$, then a new candidate axiom is generated where $c_1$ is a subclass of $c_2$, i.e., $c_1 \sqsubseteq c_2$. Each candidate axiom together with the distance $distance_{c_1c_2}$, are stored in a list. As a final axiom, we select the axiom with the smallest distance.

In the final step, the algorithm computes the transitive closure over all the previously extracted axioms (lines 25-44). We also make sure that there are no cycles in the final class hierarchy (line 34).

\begin{algorithm2e}
\SetLine
	\KwData{$KB$: Knowledge base, $V_{KB}$: Knowledge base embeddings for the knowledge base KB}
	\KwResult{$A$: Set of class subsumption axioms}
	$A$ := $\emptyset$\\
	\# Calculate the centroid and radius for each class in the knowledge base\\
	$C_{KB}=\{c \;| \; \exists i \; typeOf \; c \wedge i \in KB\}$\\
	\ForEach{class $c \in C_{KB}$}{
		$I_c = \{i\;|\;i \; typeOf \; c \wedge i \in KB \wedge \; \exists v_i \in V_{KB}\}$\\
		$c.centroid = \frac{1}{|I_c|}\sum_{i \in I_c}{v_i}$\\
		$c.radius = \sqrt{\frac{1}{|I_c|}\sum_{i \in I_c}{(v_i-c.centroid)^2}}$\\
	}
	
	\# Extract class subsumption axioms\\
	\ForEach{class $c_1 \in C$}{
		$A_{c_1}$ := $\emptyset$\\
		\ForEach{class $c_2 \in C$}{
			\If{$c_1 == c_2$}{
				continue\\
			}
			
			$distance_{c_1c_2} = distance(c_1.centroid,\; c_2.centroid)$\\
			\If{$distance_{c_1c_2} \le c_2.radius \; and \; c_1.radius < c_2.radius$}{
				$axiom = c_1 \sqsubseteq c_2$\\
				add $[axiom,distance]$ to $A_{c_1}$\\
			}
		}
		sort $A_{c_1}$ in ascending order\\
		add $A_{c_1}.first()$ to $A$
	}
	
	\# Compute transitive closure\\
	$change = true$\\
	\While{$change == true$}{ 
	\ForEach{axiom $a \in A$}{
		$change = false$\\
		$subClass = axiom.getSubClass()$\\
		$superClass = axiom.getSuperClass()$\\
		$superClasses = \{c \;|\; \exists axiom \in A \wedge axiom\; =\; superClass \sqsubseteq c\}$\\
		\ForEach{class $s \in superClasses$}{
			\If{$s == subClass$}
				{
					continue\\
				}
				$axiom = subClass \sqsubseteq s$\\
				\If{$axiom \notin A$}{
					add $axiom$ to $A$\\
						$change = true$\\
					}
		}
	}
	}
	return $A$\\
	\caption{Algorithm for class subsumption axioms extraction from a knowledge base}
	\label{algo:clusteringAlgorithm}
\end{algorithm2e}

\section{Experiments}
\label{sec:experiments}
We perform two experiments: (i) applying the proposed approach on the DBpedia dataset using the DBpedia ontology as a gold standard (see Section \ref{sec:dbpediaontoembed}); (ii) applying the proposed approach on the WebIsA database in unsupervised manner (see Section \ref{sec:webisadb}).
\label{sec:webisadb}

\subsection{Embedding the DBpedia Ontology}
\label{sec:dbpediaontoembed}

\begin{table}[t]
  \centering
  \caption{Results for class subsumption axioms extraction using DBpedia ontology as a gold standard}
    \begin{tabular}{|l|c|c|c|}
   \hline
     Method     & Precision     & Recall     & F-score \\
    \hline
    rel out & 0.056 & 0.076 & 0.064 \\ \hline
    rel in & 0.097 & 0.209 & 0.132 \\ \hline
    rel in \& out & 0.104 & 0.219 & 0.141 \\ \hline
		 \hline
    \approach & \textbf{0.594} & \textbf{0.465} & \textbf{0.521} \\ \hline
    
    \end{tabular}%
  \label{tab:dbpediaResults}%
\end{table}%

DBpedia is a public knowledge graph which is derived from structured information in Wikipedia, mainly infoboxes. For every Wikipedia page, a node in the DBpedia knowledge graph is generated, and the links to other Wikipedia pages contained in the infoboxes are transformed into labeled edges in the knowledge graph connecting the entities \cite{LehmannDBpedia}.

Besides the nodes (i.e., instances) and their interconnections, DBpedia also has an ontology, which is curated manually, and, hence, rather small. In our first experiment, we analyze how our approach is able to reconstruct the manually created ontology using the knowledge graph.

While there have been proposed multiple approaches for knowledge graph entity embeddings \cite{nickel2016review}, like RESCAL \cite{nickel2011three}, Neural Tensor Networks (NTN) \cite{socher2013reasoning}, TransE \cite{bordes2013translating}, TransH \cite{wang2014knowledge}, and TransR \cite{lin2015learning}, in this work, we focus on the RDF2vec approach \cite{ristoski2016rdf2vec}, which has been shown to provide high quality graph embeddings which can be effectively used for calculating similarity and relatedness between entities. 

The RDF2vec approach works in two phases: (i) transform the graph into a set of sequences of entities; (ii) feed the sequences of entities into a neural language models. This results into vector representation of all the nodes in the graph in a latent feature space, where similar entities appear close to each other. 

To generate the entity embeddings we use the English version of the 2016-04 DBpedia dataset and the corresponding ontology, which contains $4,678,230$ instances (i.e., nodes in the graph) and $1,379$ mapping-based properties (i.e., types of edges in the graph). In our evaluation we only consider object properties, and ignore datatype properties and literals. When building the model we do not use the DBpedia ontology, i.e., we do not use any class subsumption information. Furthermore, we only use the instance types, and ignore the instance transitive types. To generate the gold standard, we select all classes from the ontology that have at least one instance in the knowledge graph, resulting in $415$ classes, for which the Ontology defines $632$ atomic class subsumption axioms.

To build the RDF2vec model, first we generate $250$ random walks for each entity. The complete set of sequences is later used to train a Skip-Gram model with the following parameters: window size = 5; number of iterations = 5; negative sampling for optimization; negative samples = 25; dimensions = 200; The parameters are selected based on recommendations from the literature.

We compare the embedding vectors to three baseline feature generation approaches, as proposed in \cite{PaulheimFeGeLOD}. We generate features derived from generic relations in the graph, i.e., we generate a  feature for each incoming (rel in) or outgoing relation (rel out) of an entity, ignoring the value or target entity of the relation. Furthermore, we combine both incoming and outgoing relations (rel in \& out).

The results are shown in Table \ref{tab:dbpediaResults}.
The results show that we are able to identify 46.5\% of all the class subsumption axioms, and from the ones we discovered, 59.4\% exist in DBpedia. Furthermore, the results show that using entity embeddings significantly outperforms the baseline feature generation approaches.

The error analysis showed that the algorithm often extracts subsumption axioms for classes on the same level in the hierarchy, or siblings classes, e.g. \textit{dbo:Bird $\subseteq$ dbo:Mammal}. The reason for such false positives is that the centroids of the classes are positioned very close to each other in the embeddings space. 
Furthermore, some of the false positives would not necessarily be incorrect, but those axioms simply do not exist in the DBpedia ontology, e.g., \textit{dbo:Senator $\subseteq$ dbo:OfficeHolder}.
Since the DBpedia ontology, like all data sets on the Semantic Web, follows the open world assumption, 59.4\% are only a pessimistic estimate for our approach's precision (see \cite{paulheim2017kg} for a discussion).

As a second comparison, we tried to compare our approach to the approach proposed by V{\"o}lker and Niepert \cite{volker2011statistical}, who propose to use association rule mining for deriving subsumption axioms. The core idea of their approach is that if instances of class A are often also of class B, then A should be a subclass of B. However, the approach was not able to generate any axioms on the dataset we used, since in this dataset, only the most specific class is assigned to each instance, hence, co-occurrences such as those which their approach tries to exploit are rarely found in the dataset.

\begin{table*}[hbtp]
  \centering
  \caption{Results for the class subsumption axiom induction in the \textit{Person} domain}
    \begin{tabular}{|l|c|c|c|}
    \hline
    Method & DBpedia Coverage & Extra Coverage(\%) & Precision (random 100) \\
    \hline
    Association Rules &\textbf{ 0.44}  & \textbf{31301.63} & 0.25 \\
		\hline
    RDF2Vec \approach & 0.31  & 3594.44 & \textbf{0.42} \\
		\hline
    GloVe \approach & 0.29  & 1520.00 & 0.29 \\
    \hline
    \end{tabular}%
  \label{tab:personRes}%
\end{table*}%

\begin{table*}[hbtp]
  \centering
  \caption{Results for the class subsumption axiom induction in the \textit{Place} domain}
    \begin{tabular}{|l|c|c|c|}
    \hline
    Method & DBpedia Coverage & Extra Coverage(\%) & Precision (random 100) \\
    \hline
    Association Rules & \textbf{0.25}  & \textbf{11029.71} & 0.45 \\\hline
    RDF2Vec \approach & 0.21  & 3808.57 & \textbf{0.54} \\\hline
    GloVe \approach & 0.17  & 301.14 & 0.48 \\\hline
        \end{tabular}%
  \label{tab:placeRes}%
\end{table*}%

\begin{figure*}[t]
    \centering
      \frame{\includegraphics[width=0.9\textwidth]{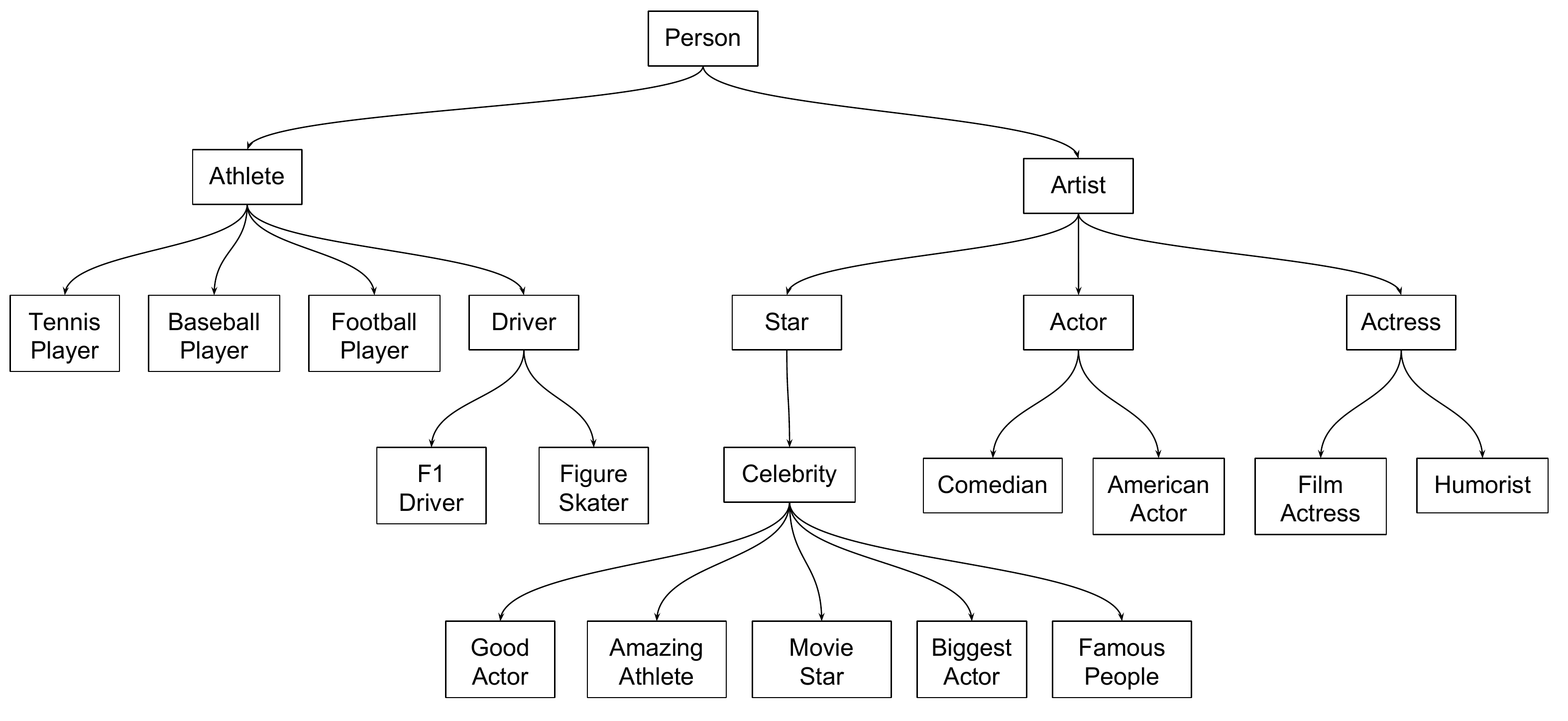}}
  \caption{Excerpt of the \textit{Person} top level hierarchy}
  \label{fig:personExc}
\end{figure*}

\subsection{Inducing Ontologies from the WebIsADb}
The WebIsA database (WebIsADb) is a database of word pairs which reflect subsumption relations. It was generated from the CommonCrawl by applying a set of patterns like Hearst patterns \cite{hearst1992automatic}, e.g., \emph{X is a Y} or \emph{Y such as X} to the large corpus of text in the common crawl. Given that the input is raw text from the Web, and all the patterns are heuristic, the corpus is assumed to provide high coverage, but low precision. Thus, simply arranging all the axioms from the WebIsADb into a graph would not constitute a taxonomy, but rather a densely connected graph with many cycles.

In the second set of experiments, we apply the proposed approach on the WebIsADb in order to extract meaningful class hierarchies for the underlying data. As the WebIsADb contains $120,992,248$ entities, extracting a single class hierarchy for the complete database is not a trivial task. Therefore, we narrowed the experiments to extract 2 domain specific class hierarchies, i.e., \textit{Person} and \textit{Place}.

To extract the class hierarchy, first we need to select the domain specific instances from the WebIsADb and identify a finite set of classes for which we need to extract class subsumption axioms. To do so, we use domain specific classes from DBpedia as filters. First, we select all the subclasses of dbo:Person (184 in total) and dbo:Place (176 in total) in DBpedia, and use them as the initial set of classes for each domain separately. Then, we select all the instances of these classes in WebIsADb. To identify the rest of the domain specific classes in WebIsADb, we expand the set of classes by adding all siblings of the corresponding class within the WebIsADb. For example, we use dbo:SoccerPlayer to select all the instances of type ``soccer player'' in WebIsADb, e.g., ``Cristiano Ronaldo'' is such an instance. In the next step, we expand the initial set of classes with all the classes assigned to the ``soccer player'' instances, e.g., from the instance ``Cristiano Ronaldo'', we will add the following classes in the set of \textit{Person} classes: ``Player'', ``Portuguese Footballer'', ``Star'', ``Great Player'', etc. 

Once we have defined the set of classes for each domain, we select all the instances for each class, which represents the input knowledge base for Algorithm \ref{algo:clusteringAlgorithm}. We experiment with entity and word embeddings. As entity embeddings we use DBpedia RDF2vec embeddings, which are built similarly as described in the previous section, only this time we also use the instance transitive types. To link the WebIsADb instances to DBpedia, we use exact string matching.

For word embeddings, we use GloVe embeddings \cite{pennington2014glove}, trained on the complete Common Crawl.\footnote{\url{https://nlp.stanford.edu/projects/glove/}} The model provides 300-dimensional embedding vectors for 2.2 million tokens. In case of multiple tokens in the WebIsADb instance, the final vector is calculated as the average of the vectors of all the tokens in the instance.

Again, we compare our approach to the association rule mining-based approach proposed in \cite{volker2011statistical}. To do so, for each instance, we generate a transaction of all the instance's types. Then, for each of the classes separately, we learn class subsumption axioms using the standard Apriori algorithm \cite{agrawal1994fast}. We consider all the rules with support and confidence value above 50.

To evaluate the induced class subsumption axioms, we use the DBpedia ontology as a reference class hierarchy, i.e., (i) we count how many of the class subsumption axioms defined in the DBpedia ontology we identified in WebIsADb (DBpedia coverage); (ii) we count how many more axioms were discovered compared to the DBpedia Ontology (Extra Coverage); (iii) we manually determine the precision on 100 randomly selected axioms from all of the extracted axioms.

First, we manually map each DBpedia class to the corresponding WebIsADb class. For the \textit{Person} domain, we were able to map 0.99\% of the classes, and there is 1466.67\% extra class coverage for the same domain in WebIsADb. For the \textit{Place} domain, we were able to map 0.71\% of the classes, and there is 1161.93\% extra class coverage for the same domain in WebIsADb.
The results for the \textit{Person} and the \textit{Place} domain are shown in Table \ref{tab:personRes} and \ref{tab:placeRes}, respectively. We can observe that although the coverage of axioms is slightly lower using the graph embeddings and \approach, we are able to mine subsumption axioms at much higher precision.

\begin{figure*}[t]
    \centering
      \frame{\includegraphics[width=0.9\textwidth]{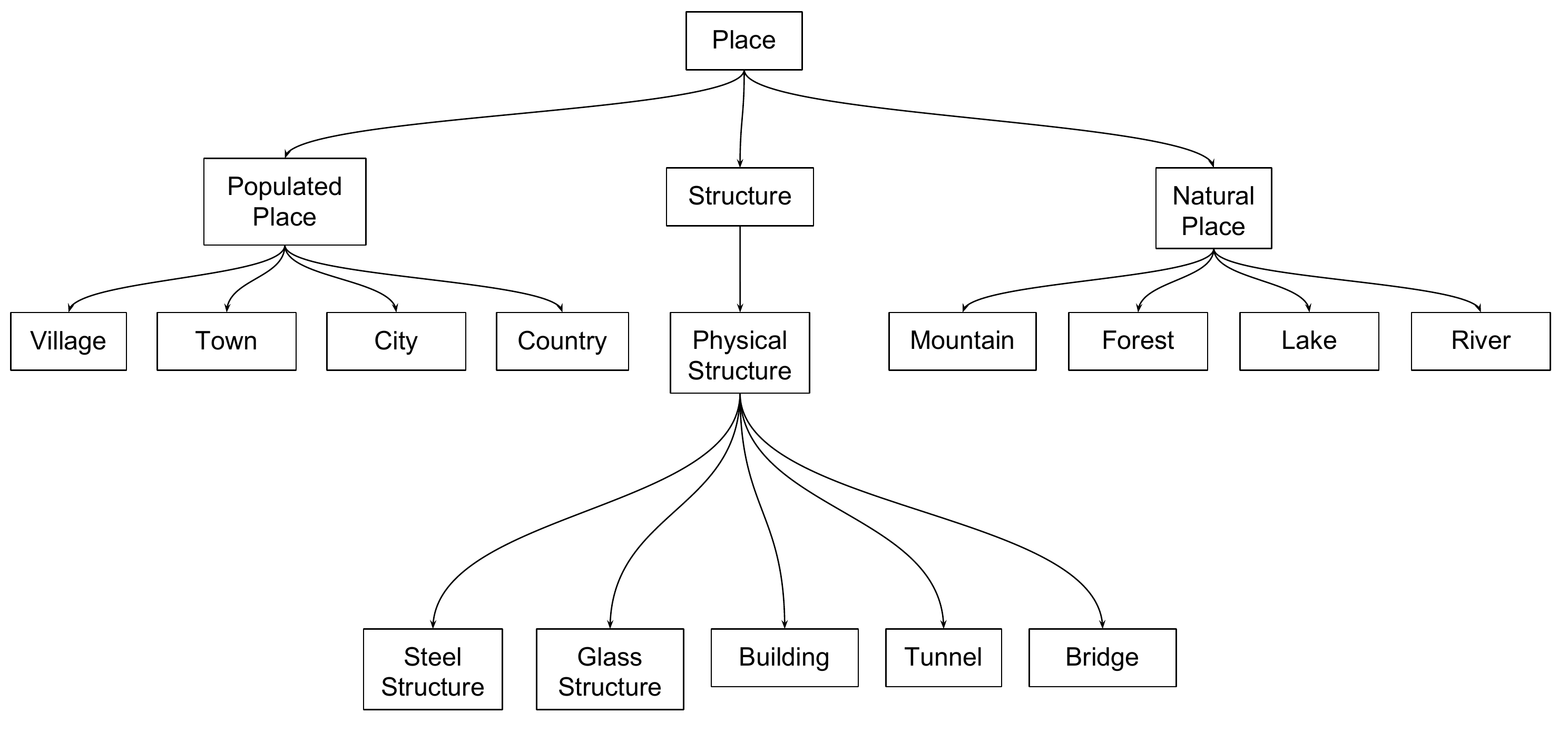}}
  \caption{Excerpt of the \textit{Place} top level hierarchy}
  \label{fig:palceExc}
\end{figure*}

The extracted class subsumption axioms can be found online.\footnote{\url{http://data.dws.informatik.uni-mannheim.de/rdf2vec/TI/webisa/}} Excerpts of the top levels of the \textit{Person} and \textit{Place} hierarchies are shown in Figure \ref{fig:personExc} and Figure \ref{fig:palceExc}, respectively.

\section{Conclusion and Outlook}
\label{sec:conclusion}
Taxonomies are an important backbone of formal knowledge representations. However, at large scale, they cannot be created manually with reasonable efforts, thus, automatic approaches have to be used.
In this paper, we have accordingly shown that using word and knowledge base embeddings are suitable approaches for inducing large scale taxonomies from knowledge graphs such as DBpedia or the WebIsADB. The approach relies on vector space embeddings of entities and exploits the proximity preserving properties of such embeddings approaches such as GloVe or RDF2vec.

Using WebIsADB, we were able to create hierarchies of thousands of classes at decent precision. We created two example hierarchies, i.e., persons and places, but the approach is capable of generating class hierarchies for any seed concept at hand. In general, the approach cannot only be used for taxonomy induction, but also for the problem of type prediction \cite{paulheim2013type}. Here, again exploiting the proximity relations in the embedding space, each instance can be assigned to the types with the closest cluster centroid(s).

In the future, it will be interesting to see how embeddings coming from text and graphs can be combined reasonably. This will allow for even more concise induction of taxonomies. Furthermore, we want to investigate how higher-level semantic knowledge, such as class restrictions or complementarity, can be mined using an embedding-based approach. That way, using the Common Crawl as a representative sample of the knowledge that exists on the Web, we will be able to create large-scale semantic knowledge representations directly from Web data.

\section*{Acknowledgments}
The work presented in this paper has been partially funded by the Junior-professor funding programme of the Ministry of Science, Research and the Arts of the state of Baden-W{\"u}rttemberg (project ``Deep semantic models for high-end NLP application''), and by the German Research Foundation (DFG) under grant number PA 2373/1-1 (Mine@LOD). 

\balance
\bibliographystyle{plain}
\bibliography{ref}

\end{document}